\documentclass[11pt]{article}
\usepackage{amsmath}
\usepackage{algorithm}

\usepackage{acl}

\usepackage{amsmath}
\usepackage{amssymb}
\usepackage{makecell}
\usepackage{mathtools}
\usepackage{amsthm}
\usepackage{multirow}
\usepackage{times}
\usepackage{cleveref}
\usepackage{latexsym}
\usepackage{booktabs}
\usepackage{algorithmicx}
\usepackage{booktabs}
\usepackage{makecell}
\usepackage[table]{xcolor}
\usepackage{pifont}
\usepackage{graphicx}
\usepackage{subcaption}

\definecolor{coseblue}{RGB}{232,242,255}
\usepackage[T1]{fontenc}

\usepackage[utf8]{inputenc}

\usepackage{microtype}

\usepackage{inconsolata}

\usepackage{graphicx}

\usepackage[table]{xcolor}
\usepackage{colortbl}
\usepackage{subcaption}
\definecolor{t1head}{RGB}{231,240,248}   
\definecolor{t1group}{RGB}{244,247,250}  
\definecolor{t1cose}{RGB}{224,236,245}   
\definecolor{t2head}{RGB}{249,242,229}   
\definecolor{t2group}{RGB}{251,248,242}  
\definecolor{t2cose}{RGB}{246,236,221}   
\definecolor{t3head}{RGB}{231,243,239}   
\definecolor{t3group}{RGB}{245,249,247}  
\definecolor{t3cose}{RGB}{223,238,232}   

\title{Confidence-Orchestrated Self-Evolution against Uncertain LLM Feedback}

\author{Bowen Wei \\
  George Mason University \\
  \texttt{bwei2@gmu.edu} \\\And
  Nan Wang \\
  Amazon \\
  \texttt{nanww@amazon.com} \\\And
  Yuqing Zhou \\
  George Mason University \\
  \texttt{yzhou31@gmu.edu} \\\AND
  Jinhao Pan \\
  George Mason University \\
  \texttt{jpan23@gmu.edu} \\\And
  Ziwei Zhu \\
  George Mason University \\
  \texttt{zzhu20@gmu.edu} \\}

\begin{document}
\maketitle
\begin{abstract}
Self-evolving large language models (LLMs) learn by generating their own training tasks and solutions, reducing reliance on human-curated supervision. However, in many reasoning domains, the model must also validate generated tasks and judge generated answers to obtain training signals. This creates a training-signal challenge: erroneous self-judgments become erroneous gradient updates. Existing approaches either rely on external verifiers, which limits generality, or treat noisy self-generated feedback as supervision. We propose COSE (Confidence-Orchestrated Self-Evolution), which uses the LLM's intrinsic confidence as a lightweight uncertainty signal to modulate learning. COSE introduces confidence-weighted PPO updates and confidence-prioritized replay. Across 19 held-out benchmarks and four Qwen/Llama backbones (0.6B--4B), COSE consistently improves over base models and achieves the best average performance in general reasoning and mathematics, while remaining competitive on code. Code and data are available at \url{https://anonymous.4open.science/r/COSE_-B5C2}.
\end{abstract}

\section{Introduction}
\label{sec:intro}

Self-evolving LLMs aim to improve reasoning ability without relying entirely on human-curated supervision. A model proposes training tasks, validates them, solves them, judges the resulting answers, and learns from the feedback~\citep{zhao2025absolute,huang2025rzero,liu2025spice}. This paradigm is attractive because high-quality reasoning data is difficult to scale: many tasks require expert-written problems, reference answers, or domain-specific annotation~\citep{villalobos2024data}. Recent progress in reinforcement learning with verifiable rewards further shows that reliable feedback can substantially improve LLM reasoning~\citep{guo2025deepseek,jaech2024openai}; self-evolution seeks to obtain such feedback automatically.

\begin{table*}[t]
\centering
\small
\setlength{\tabcolsep}{5pt}
\renewcommand{\arraystretch}{1.12}
\begin{tabular}{l|l|l|l}
\toprule
\textbf{Paradigm}
& \textbf{Training data}
& \textbf{Feedback source}
& \textbf{Limitation} \\
\midrule
SFT~\citep{radford2018improving}     
& Human Q\&A           
& Human label         
& Annotation does not scale. \\

RLVR~\citep{guo2025deepseek}
& Human Q, self A      
& Verifier            
& Human tasks still required. \\

AZR~\citep{zhao2025absolute}
& Self-generated Q\&A  
& Python verifier     
& Executable domains only. \\

MAE~\citep{chen2025mae}
& Self-generated Q\&A  
& LLM Judge           
& Judge errors enter gradients. \\

R-Zero~\citep{huang2025rzero}
& Self-generated Q\&A  
& Majority-vote LLM Judge 
& No per-sample confidence control. \\

\rowcolor{coseblue}
\textbf{COSE}
& \textbf{Self-generated Q\&A} 
& \textbf{LLM Judge with confidence} 
&  \\
\bottomrule
\end{tabular}
\caption{Roadmap from human supervision to confidence-modulated self-evolution.
Each row removes a source of supervision until self-judging methods hit a new
bottleneck: noisy LLM-generated feedback. COSE addresses this by modulating each
update using the model's confidence while producing the feedback.}
\label{tab:method_comparison}
\end{table*}

The central bottleneck is noisy self-generated feedback. In domains such as code and some mathematics, generated solutions can often be checked by execution, exact matching, or other verifiable mechanisms. Outside these domains, self-evolving systems must rely on LLM-generated feedback: a Validator decides whether a generated task is valid, and a Judge decides whether a generated answer is correct. These signals are fragile. A model that cannot reliably solve a difficult problem may also fail to recognize whether the problem is well-formed or whether a proposed solution is correct~\citep{stechly2024selfverification,huang2024cannotcorrect,panickssery2024selfpreference}.

This uncertainty is not merely an evaluation issue; it becomes an optimization issue. In self-evolution, validation and judging decisions are converted into rewards, replay priorities, or filtering decisions. False endorsements reward flawed reasoning, while false rejections discard useful supervision. Over repeated self-evolution cycles, these errors can be replayed and amplified by policy updates. The key question is therefore not only how to generate more self-training data, but how to control the training impact of uncertain LLM-generated feedback.

We propose \textbf{COSE} (\textbf{Co}nfidence-\textbf{O}rchestrated \textbf{S}elf-\textbf{E}volution), a confidence-aware framework for self-evolving reasoning training. Its core intuition is simple: \emph{do not treat all LLM-generated feedback equally}. COSE estimates the confidence of Validator and Judge feedback from the entropy of the model's token-level output distribution, following prior work on entropy-based confidence estimation~\citep{geifman2017selective,malinin2018predictive,geng2024survey}.This confidence is not a correctness guarantee; instead, it provides a lightweight signal for reducing the influence of uncertain feedback. COSE uses this signal in two places: confidence-weighted PPO, which reduces the gradient contribution of uncertain feedback, and confidence-prioritized replay, which samples more from confidently validated questions. In this way, COSE turns confidence into a training-control signal for self-evolution.

Our contributions are threefold:
\begin{itemize}
    \item We identify noisy LLM-generated feedback as a central optimization bottleneck in self-evolving LLMs, covering both validation of generated tasks and judgment of generated answers.

    \item We formulate noisy LLM-generated feedback as a per-sample training-control problem in self-evolution, where each validation or judgment signal affects learning according to its confidence.

    \item We instantiate this formulation through confidence-weighted PPO and confidence-prioritized replay, and show across 19 held-out benchmarks and four Qwen/Llama backbones that COSE consistently improves over base models and outperforms or remains competitive with prior self-evolving baselines. Code and data are available at \url{https://anonymous.4open.science/r/COSE_-B5C2}.
\end{itemize}

\section{Related Work}
\label{sec:related}

Table~\ref{tab:method_comparison} situates COSE within the progression from human supervision to self-evolving training. SFT provides reliable human labels but is difficult to scale; RLVR replaces human answer labels with verifiable rewards, but still depends on human-written tasks and verifiable domains~\citep{guo2025deepseek,jaech2024openai}; AZR removes human-written tasks through self-generated examples, but remains tied to Python-executable verification~\citep{zhao2025absolute}; and MAE/R-Zero extend self-evolution to open-domain reasoning with LLM Judges, but largely treat accepted feedback as uniformly weighted training signal~\citep{chen2025mae,huang2025rzero}. COSE addresses this missing step by keeping self-generated tasks and open-domain LLM feedback while adding per-sample training control through confidence-weighted PPO and confidence-prioritized replay. Detailed related work on self-evolving LLMs, LLM-as-Judge uncertainty, confidence estimation, and sample weighting is provided in Appendix~\ref{app:related}.

\begin{figure*}[t]
\centering
\includegraphics[width=0.9\textwidth]{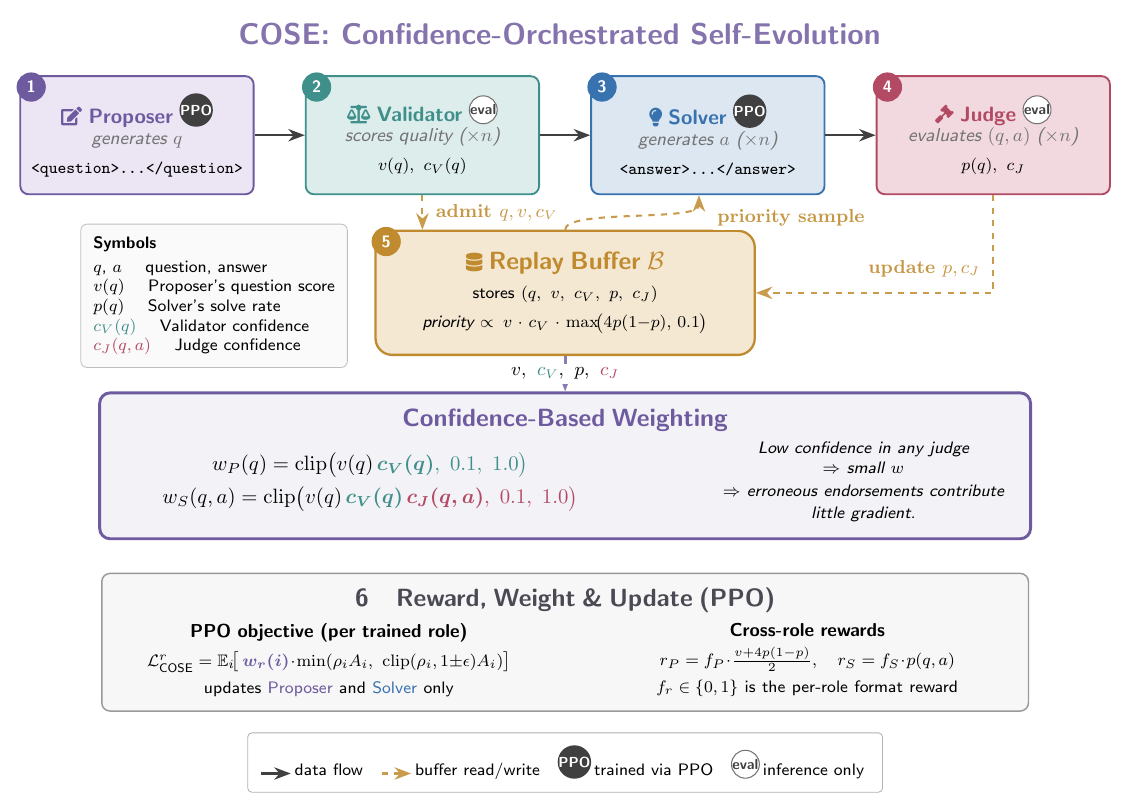}
\caption{Overview of COSE. A single language model $\pi_\theta$ is prompted to play four roles: Proposer, Validator, Solver, and Judge. The Proposer generates candidate questions, the Validator scores and filters them, the Solver answers replayed questions, and the Judge scores answer correctness. COSE computes sequence-level confidence estimates from token-level normalized entropy and uses them to regulate training through confidence-weighted PPO updates and confidence-prioritized replay.}
\label{fig:architecture}
\vspace{-10pt}
\end{figure*}

\section{Method}
\label{sec:method}

We introduce \textbf{COSE} (Confidence-Orchestrated Self-Evolution), a self-evolving framework that uses confidence from LLM-generated feedback to make training more robust. COSE follows a simple principle: feedback from a Validator or Judge should influence learning according to the model's confidence in producing that feedback. Figure~\ref{fig:architecture} gives an overview. COSE uses a single backbone model $\pi_\theta$ with four prompt-conditioned roles: \emph{Proposer}, \emph{Validator}, \emph{Solver}, and \emph{Judge}. The Proposer generates candidate reasoning questions, the Validator checks whether those questions are valid and useful, the Solver produces answers, and the Judge evaluates answer correctness. During training, only Proposer and Solver trajectories receive optimization losses. The Validator and Judge are used as inference-time feedback modules that provide scores and confidence. 

\paragraph{Notation.}
We use $q$ for a generated question and $a$ for a Solver answer. The Validator assigns each question a normalized quality score $v(q)\in[0,1]$ and a confidence score $c_V(q)\in[0,1]$. The Judge assigns each answer a normalized correctness score $p(q,a)\in[0,1]$ and a confidence score $c_J(q,a)\in[0,1]$. A score of $0$ denotes invalid or incorrect feedback, while a score of $1$ denotes fully valid or correct feedback. If the Validator or Judge emits a discrete score, we linearly normalize it to $[0,1]$ before computing rewards, replay priorities, or PPO weights. For replay, we maintain a question-level solve rate:
\begin{equation}
p(q)=\frac{1}{|\mathcal{A}_q|}\sum_{a\in\mathcal{A}_q}p(q,a),
\end{equation}
where $\mathcal{A}_q$ is the set of sampled Solver answers for question $q$. Thus, $p(q)\in[0,1]$ and is used for question-level replay priority and Proposer rewards, while $p(q,a)$ is used for Solver rewards.

\subsection{Self-Evolution Loop}
\label{sec:workflow}

Each training iteration has two stages. First, the Proposer generates candidate questions. The Validator scores each question and filters out invalid or low-quality ones. Accepted questions are stored in a replay buffer $\mathcal{B}$ together with their validation score $v(q)$ and Validator confidence $c_V(q)$. The Validator is used only as an inference-time feedback module for this step; no PPO loss is applied to Validator trajectories.

Second, the Solver samples questions from $\mathcal{B}$ and generates answers. For each answer $a$ to question $q$, the Judge assigns a correctness score $p(q,a)$ and confidence $c_J(q,a)$. When multiple Solver answers are generated for the same question, COSE updates the question-level solve rate $p(q)$ by averaging their Judge scores. The replay buffer is updated with $p(q)$, allowing future sampling to prefer questions that are valid, confidently judged, and near the Solver's current capability boundary. Like the Validator, the Judge provides inference-time feedback only; no PPO loss is applied to Judge.

Although all four roles share the same backbone checkpoint, COSE applies optimization only through the Proposer and Solver roles. The Validator and Judge influence training through their scores and confidence estimates, but they are not directly optimized as objectives. This separation is important: COSE does not assume that LLM feedback is always correct, but instead controls how strongly each feedback signal affects learning.

\begin{table*}[t]
\centering
\small
\setlength{\tabcolsep}{3.5pt}
\begin{tabular}{l|cccccccccc|c}
\toprule
\rowcolor{white} & \multicolumn{10}{c|}{\textbf{General Reasoning}} & \\
\rowcolor{t1head}
\textbf{Method} & \textbf{MMLU} & \textbf{GPQA} & \textbf{BBH} & \textbf{LB-R} & \textbf{ARC-C} & \textbf{CQA} & \textbf{OBQA} & \textbf{NQ} & \textbf{TriviaQA} & \textbf{SQuAD} & \textbf{Avg.} \\
\midrule
\rowcolor{t1group} \multicolumn{12}{c}{\textit{Qwen3-0.6B}} \\
\midrule
Base          & 44.00 & 30.00 & 51.00 & 26.00 & 68.00 & 55.00 & 73.00 & 16.00 & 18.00 & 85.00 & 46.60 \\
AZR           & 52.40 & 32.23 & 53.06 & 28.22 & 69.80 & 57.63 & 73.40 & 18.04 & 21.75 & 89.21 & 49.57 \\
MAE           & 55.20 & 35.10 & 56.80 & 31.40 & 72.30 & 61.20 & 76.40 & 20.50 & 25.80 & 89.60 & 52.43 \\
R-Zero        & 53.60 & 34.80 & 55.40 & 30.20 & 71.40 & 60.10 & 75.20 & 19.80 & 24.60 & 88.90 & 51.40 \\
\rowcolor{t1cose} \textbf{COSE} & \textbf{95.00} & \textbf{82.67} & \textbf{91.85} & \textbf{97.00} & \textbf{97.00} & \textbf{98.86} & \textbf{95.96} & \textbf{39.75} & \textbf{71.55} & \textbf{95.33} & \textbf{86.50} \\
\midrule
\rowcolor{t1group} \multicolumn{12}{c}{\textit{Qwen2.5-3B-Instruct}} \\
\midrule
Base          & 63.40 & 34.67 & 53.79 & 20.80 & 80.60 & 66.80 & 67.80 & 27.80 & 51.60 & 78.20 & 54.55 \\
AZR           & 63.40 & \textbf{38.93} & 52.57 & 22.27 & 82.80 & 74.36 & 77.55 & 35.33 & 56.67 & 90.85 & 59.47 \\
MAE           & 61.40 & 35.42 & 57.51 & 23.48 & \textbf{84.20} & 71.54 & 79.39 & 37.98 & \textbf{60.13} & 92.28 & 60.33 \\
R-Zero        & 62.80 & 35.60 & 55.20 & 22.90 & 81.90 & 72.40 & 76.80 & 36.10 & 58.20 & 91.40 & 59.33 \\
\rowcolor{t1cose} \textbf{COSE} & \textbf{67.50} & 34.67 & \textbf{58.89} & \textbf{28.80} & 83.00 & \textbf{78.33} & \textbf{79.67} & \textbf{45.67} & 53.00 & \textbf{95.01} & \textbf{62.45} \\
\midrule
\rowcolor{t1group} \multicolumn{12}{c}{\textit{Llama-3.2-3B-Instruct}} \\
\midrule
Base          & 54.00 & 38.00 & 62.00 & 32.00 & 67.00 & 64.00 & 79.00 & 27.00 & 58.00 & 72.00 & 55.30 \\
AZR           & 57.60 & 39.73 & 62.72 & 34.12 & 73.40 & 68.73 & 80.32 & 37.26 & 58.23 & 83.46 & 59.56 \\
MAE           & \textbf{95.00} & 81.00 & 88.00 & 93.00 & \textbf{98.00} & 98.00 & 96.00 & 35.00 & 59.00 & 92.00 & 83.50 \\
R-Zero        & 58.20 & 42.60 & 64.92 & 36.56 & 73.20 & 70.27 & 81.39 & 32.78 & 59.53 & 78.18 & 59.76 \\
\rowcolor{t1cose} \textbf{COSE} & 94.80 & \textbf{83.02} & \textbf{91.51} & \textbf{93.51} & 96.80 & \textbf{98.35} & \textbf{97.67} & \textbf{41.06} & \textbf{71.86} & \textbf{94.52} & \textbf{86.31} \\
\midrule
\rowcolor{t1group} \multicolumn{12}{c}{\textit{Qwen3-4B}} \\
\midrule
Base          & 93.00 & \textbf{85.00} & \textbf{92.00} & 75.00 & 97.00 & 97.00 & \textbf{97.00} & 28.00 & 66.00 & 91.00 & 82.10 \\
AZR           & 93.40 & 83.20 & 90.60 & 78.80 & 96.80 & 97.40 & 96.60 & 36.20 & 68.40 & 94.20 & 83.56 \\
MAE           & 94.20 & 82.60 & 90.80 & 79.40 & 96.90 & 97.60 & 96.80 & 35.40 & 68.90 & 94.00 & 83.66 \\
R-Zero        & 93.80 & 82.40 & 90.30 & 78.10 & 96.70 & 97.30 & 96.40 & 35.80 & 68.20 & 93.80 & 83.28 \\
\rowcolor{t1cose} \textbf{COSE} & \textbf{96.20} & 84.85 & 91.29 & \textbf{94.41} & \textbf{97.40} & \textbf{98.68} & 96.65 & \textbf{40.25} & \textbf{73.33} & \textbf{95.15} & \textbf{86.82} \\
\bottomrule
\end{tabular}
\caption{General reasoning results across four base models on 10 benchmarks. Best result is in \textbf{bold}.}
\label{tab:main_results}
\vspace{-10pt}
\end{table*}

\subsection{Confidence from LLM Feedback}
\label{sec:confidence}

COSE estimates confidence directly from the token probabilities produced when the Validator or Judge generates feedback. Intuitively, if the model's output distribution is concentrated during feedback generation, the feedback is more confident; if the distribution is diffuse, it is less confident.

In our main experiments, COSE uses \emph{normalized entropy confidence}. Entropy-based confidence and uncertainty estimates are standard in selective prediction and uncertainty estimation~\citep{geifman2017selective,malinin2018predictive}, and have recently been studied for LLM generation confidence~\citep{lin2024generating,geng2024survey}. For each generated token $t$, let $P_t$ be the model's next-token distribution over vocabulary $\mathcal{V}$. We compute token-level confidence as
\begin{equation}
c_t = 1 - \frac{H(P_t)}{\log |\mathcal{V}|},
\end{equation}
where $H(P_t)$ is the entropy of $P_t$. Since $0 \leq H(P_t) \leq \log |\mathcal{V}|$, we have $c_t\in[0,1]$. A uniform distribution gives confidence $0$, while a one-hot distribution gives confidence $1$.

For a Validator or Judge output sequence with token-level confidences $\{c_t\}_{t=1}^{T}$, COSE aggregates them into a sequence-level confidence score:
\begin{equation}
c_{\mathrm{seq}}=\frac{1}{T}\sum_{t=1}^{T}c_t.
\end{equation}
This aggregation captures the model's confidence during feedback generation, rather than only its confidence in the final score token. The resulting sequence-level score becomes the Validator confidence $c_V(q)$ for generated questions and the Judge confidence $c_J(q,a)$ for judged Solver answers. Confidence is not assumed to guarantee correctness; it is used only as a lightweight signal for modulating the training impact of uncertain feedback.

\subsection{Confidence-Weighted PPO}
\label{sec:cw}

The main use of confidence is to weight PPO updates. Standard PPO gives rewards equal training weight once they are assigned, which is problematic when rewards come from LLM-generated feedback. A low-confidence judgment may be noisy or wrong, and updating strongly on such feedback can reinforce errors.

COSE therefore scales samples by confidence weights. For Proposer updates, the weight depends on question quality and Validator confidence:
\begin{equation}
w_P(q) = \mathrm{clip}\bigl(v(q)c_V(q), 0.1, 1.0\bigr).
\label{eq:wp}
\end{equation}
For Solver updates, the weight additionally includes the Judge confidence for the specific answer:
\begin{equation}
w_S(q,a) = \mathrm{clip}\bigl(v(q)c_V(q)c_J(q,a), 0.1, 1.0\bigr).
\label{eq:weights}
\end{equation}
Because $v(q)$, $c_V(q)$, and $c_J(q,a)$ are all in $[0,1]$, both weights lie in $[0.1,1.0]$. Thus, if a question is low-quality, the Validator is uncertain, or the Judge is uncertain, the corresponding update has less influence. We use a small lower bound rather than discarding uncertain samples entirely, which preserves exploration while still reducing the effect of uncertain feedback.

\subsection{Confidence-Prioritized Replay}
\label{sec:replay}

COSE also uses confidence to prioritize replay. Questions are sampled more often when they are confidently validated and have intermediate difficulty for the current Solver:
\begin{equation}
\mathrm{priority}(q) \propto v(q)c_V(q)\max(4p(q)(1-p(q)), 0.1).
\label{eq:priority}
\end{equation}
The term $4p(q)(1-p(q))$ lies in $[0,1]$ and is largest when $p(q)$ is near $0.5$, so replay focuses on questions that the Solver answers correctly about half of the time. These questions are more informative than questions that are already solved consistently or almost always failed. The factor $v(q)c_V(q)$ prevents the buffer from over-sampling questions whose validation is low-quality or uncertain.

\subsection{Rewards}
\label{sec:rewards}

COSE trains the Proposer and Solver with simple cross-role rewards. The Solver receives the Judge-assigned correctness score for its specific answer:
\begin{equation}
r_S(q,a) = f_S(a)\cdot p(q,a).
\end{equation}
The Proposer is rewarded for generating questions that are both high-quality and learnable:
\begin{equation}
r_P(q) = f_P(q)\cdot \frac{v(q) + 4p(q)(1-p(q))}{2}.
\end{equation}
Here, $f_P(q),f_S(a)\in\{0,1\}$ are format-validity indicators that ensure outputs contain the required tags. Since $v(q)$, $p(q,a)$, and $p(q)$ are normalized to $[0,1]$, both $r_S(q,a)$ and $r_P(q)$ are also in $[0,1]$. The Validator and Judge are not updated by PPO; they only provide scores and confidence estimates.
\begin{figure*}[t]
\centering
\begin{subfigure}[t]{0.32\textwidth}
\centering
\includegraphics[width=\textwidth]{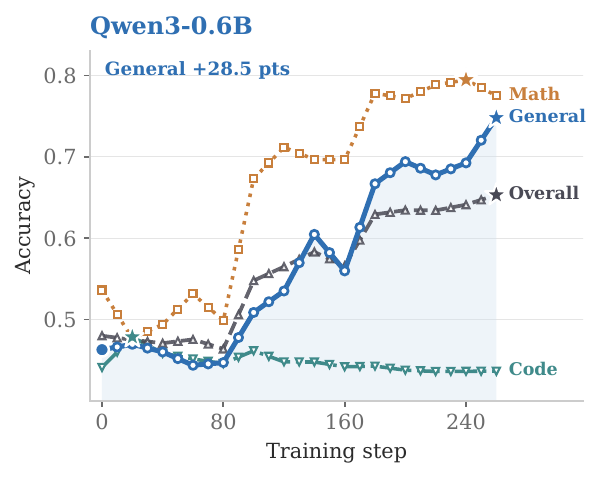}
\caption{Qwen3-0.6B}
\label{fig:curve_qwen06}
\end{subfigure}
\hfill
\begin{subfigure}[t]{0.32\textwidth}
\centering
\includegraphics[width=\textwidth]{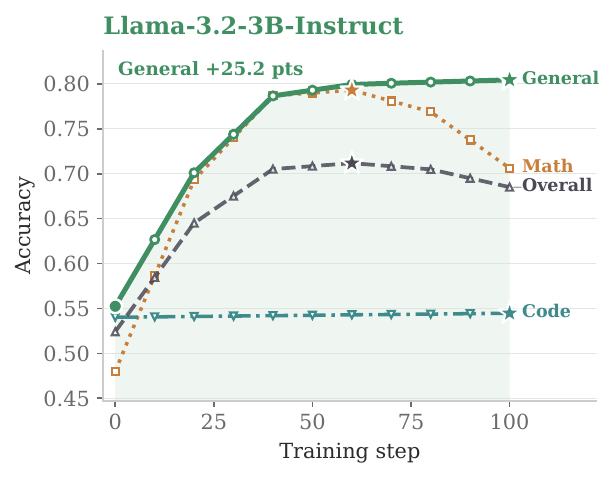}
\caption{Llama-3.2-3B-Instruct}
\label{fig:curve_llama}
\end{subfigure}
\hfill
\begin{subfigure}[t]{0.32\textwidth}
\centering
\includegraphics[width=\textwidth]{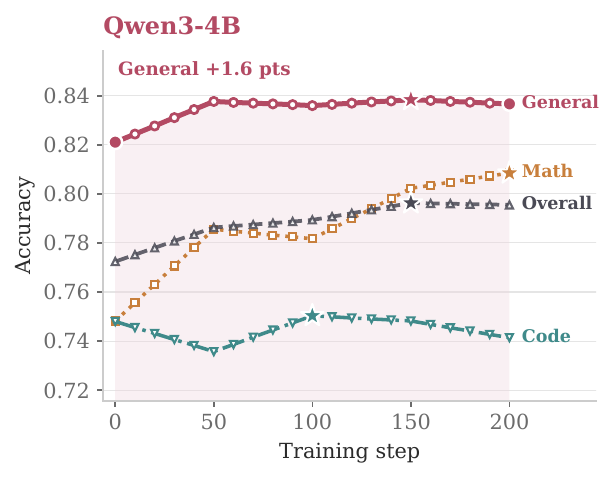}
\caption{Qwen3-4B}
\label{fig:curve_qwen4}
\end{subfigure}
\caption{Overall held-out evaluation accuracy over COSE training steps, shown separately for each base model.}
\vspace{-10pt}
\label{fig:training_curves}
\end{figure*}

\section{Experiments}
\label{sec:experiments}
We evaluate COSE on 19 held-out benchmarks across general reasoning, mathematics, and code generation. Our experiments are designed to answer seven questions:
(RQ1) Does COSE improve general reasoning?
(RQ2) Does COSE improve mathematical reasoning?
(RQ3) Does COSE transfer to code generation?
(RQ4) How stable is COSE during training?
(RQ5) Which components of COSE matter?
(RQ6) How sensitive is COSE to batch size and confidence-signal choice?
(RQ7) How does confidence weighting affect individual self-generated rollouts?

\subsection{Experimental Setup}
\label{sec:exp_setup}

\paragraph{Base models.}
We evaluate COSE on four backbones spanning two model families and a 0.6B--4B parameter range: \textbf{Qwen3-0.6B}, \textbf{Qwen2.5-3B-Instruct}, \textbf{Llama-3.2-3B-Instruct}, and \textbf{Qwen3-4B}. This set covers a small model with substantial improvement headroom, two instruction-tuned 3B-scale models, and a stronger 4B model that is closer to benchmark saturation.

\paragraph{Baselines.}
We compare COSE with the untrained \textbf{Base} model and three self-evolving baselines: \textbf{AZR}~\citep{zhao2025absolute}, which relies on a Python interpreter for verification; \textbf{MAE}~\citep{chen2025mae}, which uses in-model Judge feedback as supervision; and \textbf{R-Zero}~\citep{huang2025rzero}, which aggregates multiple model-generated outputs through self-judging. All methods are trained from the same initial checkpoints.

\paragraph{Evaluation.}
We evaluate on three benchmark groups. General reasoning includes MMLU, GPQA, BBH, LiveBench Reasoning, ARC-Challenge, CommonsenseQA, OpenBookQA, Natural Questions, TriviaQA, and SQuAD. Mathematics includes GSM8K, MATH, AMC, Minerva, and Olympiad-Bench. Code generation includes HumanEval, HumanEval+, MBPP, and MBPP+ from EvalPlus. None of these evaluation benchmarks are used during COSE training. Prompts are shown in Appendix~\ref{app:prompts}. Code and data are available at \url{https://anonymous.4open.science/r/COSE_-B5C2}.

\subsection{General Reasoning Results (RQ1)}
\label{sec:rq1_general}

Table~\ref{tab:main_results} reports general reasoning performance across four backbones and ten benchmarks. All self-evolving methods generally improve over the corresponding base models, confirming that self-generated training can provide useful supervision even without human-curated reasoning data. COSE achieves the highest average score on all four backbones, with the largest gains on Qwen3-0.6B and Llama-3.2-3B.

The model-wise pattern is informative. The strongest improvements appear on relatively weaker backbones, where the initial model has more room to improve but also produces noisier validation and judging signals. This is precisely the regime where naive self-evolution is fragile: the model can generate useful new questions, but it can also accept invalid questions or reward incorrect answers because its own feedback is noisy. AZR, MAE, and R-Zero still improve in many settings, showing that self-generated supervision is valuable, but their gains are less consistent because accepted feedback is used with nearly uniform trust. COSE improves most in this intermediate regime because it does not need the Validator or Judge to be perfectly correct; it only needs confidence to provide a useful relative signal for reducing the influence of uncertain feedback. On the strongest Qwen3-4B backbone, the gains are smaller but still positive, suggesting that confidence control remains useful even when the model starts with stronger reasoning ability and less noisy self-feedback.

\subsection{Mathematics Results (RQ2)}
\label{sec:rq2_math}

Table~\ref{tab:math_results} reports mathematics performance across five benchmarks on representative backbones, with the full four-backbone results provided in Appendix~\ref{app:math}. All self-evolving baselines improve over the corresponding base models on average, confirming that self-generated mathematical practice provides useful training signal. COSE achieves the best average performance, with especially large gains on Qwen3-0.6B and Llama-3.2-3B.

Mathematics is a stress test for self-evolution because the final answer is often short, but verifying the solution requires checking intermediate reasoning. Weaker backbones have more room to improve, but their Validators and Judges are also noisier, making naive self-evolution vulnerable to confidently accepting flawed derivations or rewarding incorrect answers. This explains why COSE gains most on Qwen3-0.6B and Llama-3.2-3B: these models benefit from additional self-generated practice, but need confidence-based training control to avoid learning too strongly from noisy feedback. On stronger backbones, the gains are smaller but still meaningful, especially on harder benchmarks such as AMC, Minerva, and Olympiad-Bench, where Judge feedback remains noisy.
\begin{table}[t]
\centering
\small
\setlength{\tabcolsep}{1pt}
\begin{tabular}{l|ccccc|c}
\toprule
\rowcolor{t3head}
\textbf{Method} & \textbf{GSM8K} & \textbf{MATH} & \textbf{AMC} & \textbf{Minerva} & \textbf{Olympiad} & \textbf{Avg.} \\
\midrule
\rowcolor{t3group} \multicolumn{7}{c}{\textit{Llama-3.2-3B-Instruct}} \\
\midrule
Base          & 63.00 & 47.00 & 27.70 & 64.10 & 38.10 & 47.98 \\
AZR           & 72.40 & 56.20 & 36.80 & 68.40 & 44.20 & 55.60 \\
MAE           & 75.20 & 59.80 & 40.60 & 70.20 & 47.60 & 58.68 \\
R-Zero        & 73.80 & 58.40 & 38.90 & 69.10 & 46.30 & 57.30 \\
\rowcolor{t3cose} \textbf{COSE} & \textbf{89.00} & \textbf{87.00} & \textbf{69.88} & \textbf{89.16} & \textbf{72.00} & \textbf{81.41} \\
\midrule
\rowcolor{t3group} \multicolumn{7}{c}{\textit{Qwen3-4B}} \\
\midrule
Base          & 85.00 & 81.00 & 62.60 & 78.70 & 66.80 & 74.82 \\
AZR           & 93.20 & 82.40 & 64.80 & 81.10 & 68.60 & 78.02 \\
MAE           & 92.80 & 83.20 & 65.60 & 81.80 & 69.40 & 78.56 \\
R-Zero        & 92.40 & 82.60 & 64.90 & 81.20 & 68.90 & 78.00 \\
\rowcolor{t3cose} \textbf{COSE} & \textbf{94.40} & \textbf{86.40} & \textbf{68.80} & \textbf{88.38} & \textbf{78.72} & \textbf{83.34} \\
\bottomrule
\end{tabular}
\caption{Math results across two base models.}
\label{tab:math_results}
\vspace{-10pt}
\end{table}

\subsection{Code Generation Results (RQ3)}
\label{sec:rq2_code}

Table~\ref{tab:cross-framework-eval} reports pass@1 on four EvalPlus code benchmarks. COSE is competitive across all backbones and obtains the best average code performance on Qwen3-0.6B and Qwen2.5-3B-Instruct. On Qwen3-4B and Llama-3.2-3B, AZR is slightly stronger, which is expected because code is the most favorable setting for AZR: Python execution provides an external verifier that is more reliable than an LLM Judge.

This result clarifies the role of COSE. COSE is not designed to replace exact verifiers when they are available; executable feedback remains the strongest signal for code-like tasks. Instead, COSE targets the broader setting where exact verification is unavailable or incomplete. The important observation is that COSE does not sacrifice code ability while improving general reasoning and mathematics. This suggests that confidence-weighted training does not merely overfit to noisy natural-language reasoning tasks; it provides a conservative update rule that can coexist with domains where feedback is already relatively reliable. In other words, COSE is most valuable when verification is weak, but it remains compatible with domains where stronger feedback exists. The full code table is reproduced in Appendix~\ref{app:code},
which discusses how the absence of regression on code closes the
alternative explanation that COSE's math gains come from a domain-specific
training instability.
\begin{table}[t]
\centering
\setlength{\tabcolsep}{2pt}
\small
\begin{tabular}{lccccc}
\toprule
\rowcolor{t2head}
Method & HumanEval & HumanEval+ & MBPP & MBPP+ & Avg. \\
\midrule
\rowcolor{t2group} \multicolumn{6}{c}{\textit{Qwen3-4B}} \\
\midrule
Base   & 78.66 & 73.17 & 79.63 & 67.72 & 74.80 \\
AZR    & \textbf{80.50} & \textbf{75.00} & 79.90 & 67.50 & \textbf{75.73} \\
R-Zero & 78.00 & 73.20 & \textbf{80.40} & \textbf{68.30} & 74.98 \\
\rowcolor{t2cose} COSE   & 79.30 & 73.80 & 79.90 & 67.50 & 75.13 \\
\midrule

\rowcolor{t2group} \multicolumn{6}{c}{\textit{Llama-3.2-3B-Instruct}} \\
\midrule
Base   & 54.88 & 50.61 & 61.11 & 49.47 & 54.02 \\
AZR    & \textbf{55.50} & \textbf{52.40} & \textbf{63.20} & \textbf{54.50} & \textbf{56.40} \\
R-Zero & 54.90 & 50.60 & 60.60 & 50.50 & 54.15 \\
\rowcolor{t2cose} COSE   & 53.70 & 49.40 & \textbf{63.20} & 51.90 & 54.55 \\

\bottomrule
\end{tabular}
\caption{Code benchmark results across two backbones. }
\label{tab:cross-framework-eval}
\vspace{-10pt}
\end{table}

\subsection{Training Dynamics (RQ4)}
\label{sec:rq4_dynamics}

Figure~\ref{fig:training_curves} shows held-out performance over COSE training. The curves reveal that improvement is not uniform across domains: mathematics and general reasoning account for most of the gains, while code is comparatively stable. This is consistent with the different feedback regimes across domains. Code has relatively concrete correctness signals and less room for confidence control to change the update direction, whereas math and general reasoning rely more heavily on LLM feedback and therefore benefit more from regulating uncertain judgments.

The curves also show that self-evolution is not simply a monotonic data-scaling process. Some models improve quickly and then plateau or fluctuate, especially on mathematics. This behavior is expected because the replay buffer gradually changes the training distribution: as the Solver improves, previously useful questions become easy, while newly generated harder questions may have noisier validation or judging. COSE reduces the effect of noisy feedback, but it does not eliminate distribution shift inside the self-evolution loop. The training dynamics therefore support a practical takeaway: confidence control improves stability, but checkpoint selection and replay balance remain important for long self-evolution runs.
A per-benchmark view is provided in
Appendix~\ref{app:dynamics}. 
\subsection{Ablation Study (RQ5)}
\label{sec:ablation}
Table~\ref{tab:ablation} studies the contribution of COSE's main mechanisms. The full model uses both confidence-weighted PPO and confidence-prioritized replay. We compare it with three variants: \emph{w/o weighting}, which removes confidence weights from PPO updates; \emph{w/o priority}, which removes confidence from replay sampling; and \emph{$1-p$ learnability}, which replaces the intermediate-difficulty replay signal $4p(q)(1-p(q))$ with the pure difficulty signal $1-p(q)$. Removing confidence-weighted PPO causes the largest drop, especially in mathematics, showing that the most damaging failure mode is not merely sampling imperfect questions, but applying full-strength gradients from uncertain feedback. Removing replay priority has a smaller but consistent effect, indicating that buffer sampling matters mainly by shaping the curriculum. The $1-p$ variant can degrade performance because always prioritizing hard questions over-samples examples that may be unsolvable, invalid, or beyond the current Solver's learning frontier. The ablation therefore separates two roles of COSE: PPO weighting protects optimization from noisy labels, while replay priority improves which questions the model practices on.

\begin{table}[t]
\centering
\small
\setlength{\tabcolsep}{1.5pt}
\begin{tabular}{l|ccc|ccc}
\toprule
& \multicolumn{3}{c|}{\textit{Qwen3-0.6B}} & \multicolumn{3}{c}{\textit{Llama-3.2-3B}} \\
\textbf{Variant} & General & Math & Code & General & Math & Code \\
\midrule
COSE (full)               & \textbf{85.36} & \textbf{79.64} & \textbf{48.33} & \textbf{86.31} & \textbf{81.41} & \textbf{54.47} \\
\midrule
w/o weighting             & 80.74 & 69.28 & 45.12 & 81.03 & 68.92 & 50.36 \\
w/o priority              & 83.12 & 75.31 & 46.87 & 84.02 & 76.44 & 52.18 \\
$1{-}p$ learnability      & 82.65 & 72.84 & 47.05 & 83.76 & 73.19 & 52.61 \\
\bottomrule
\end{tabular}
\caption{Ablation study on two backbones.}
\label{tab:ablation}
\vspace{-10pt}
\end{table}

\subsection{Hyperparameter Study (RQ6)}
\label{sec:hyperparam}

RQ6 studies whether COSE is sensitive to two practical choices: training batch size and the confidence signal used for Validator and Judge feedback. Since COSE uses confidence as a relative uncertainty signal rather than a calibrated probability of correctness, we expect moderate robustness to these choices.

\noindent\textbf{Training batch size.} Table~\ref{tab:bs_sweep} shows that COSE is not highly sensitive to batch size: all settings achieve similar downstream performance on both backbones. The mild upward trend with larger batches is nevertheless meaningful. Larger batches expose PPO to a more diverse set of self-generated questions and reduce variance in the confidence-weighted update. This is useful in self-evolution because individual feedback signals can be noisy, and small batches may overreact to a few confidently wrong or unusually difficult examples. The trend suggests that batch size mainly affects training smoothness rather than changing the qualitative behavior of COSE.

\noindent\textbf{Confidence signal.} Table~\ref{tab:signal_sweep} compares several intrinsic confidence signals on Qwen3-0.6B. All signals perform similarly, with normalized entropy slightly better overall. This suggests that COSE is not highly sensitive to the exact confidence estimator. The key requirement is not a perfectly calibrated confidence formula, but a signal that identifies uncertain feedback so that its influence on PPO updates can be reduced.

\begin{table}[t]
\centering
\small
\setlength{\tabcolsep}{2pt}
\begin{tabular}{l|ccc|ccc}
\toprule
& \multicolumn{3}{c|}{\textit{Qwen3-0.6B}} & \multicolumn{3}{c}{\textit{Llama-3.2-3B}} \\
\textbf{Batch size} & General & Math & Code & General & Math & Code \\
\midrule
$B = 32$            & 84.71 & 78.36 & 47.82 & 85.42 & 80.35 & 53.61 \\
$B = 64$            & 85.36 & 79.64 & 48.33 & 85.87 & 80.92 & 54.06 \\
$B = 128$           & 85.58 & 80.02 & 48.61 & 86.31 & 81.41 & 54.47 \\
\bottomrule
\end{tabular}
\caption{Effect of training batch size.}
\label{tab:bs_sweep}
\vspace{-10pt}
\end{table}

\begin{table}[t]
\centering
\small
\setlength{\tabcolsep}{4pt}
\begin{tabular}{l|ccc}
\toprule
\textbf{Confidence signal} & \textbf{General} & \textbf{Math} & \textbf{Code} \\
\midrule
Norm. entropy (COSE) & \textbf{85.36} & \textbf{79.64} & \textbf{48.33} \\
Self-certainty       & 85.01 & 79.18 & 48.06 \\
Margin               & 84.93 & 78.96 & 47.91 \\
Max logit            & 84.81 & 78.84 & 47.78 \\
Neg. entropy         & 85.12 & 79.27 & 48.11 \\
\bottomrule
\end{tabular}
\caption{Effect of confidence signal on Qwen3-0.6B.}
\label{tab:signal_sweep}
\vspace{-10pt}
\end{table}
\subsection{Qualitative Examples (RQ7)}
\label{sec:qualitative}

\begin{figure}[t]
\centering
\includegraphics[width=\columnwidth]{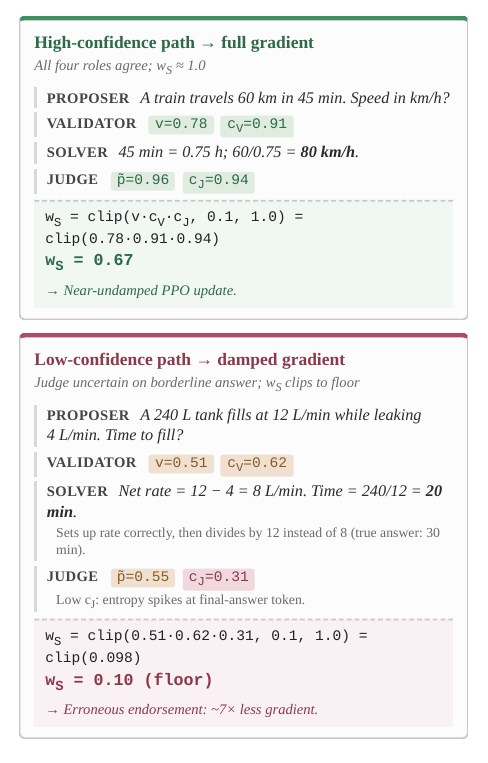}
\vspace{-20pt}
\caption{Two examples illustrating confidence-weighted gradient contribution.}
\label{fig:qualitative}
\vspace{-15pt}
\end{figure}

RQ7 illustrates how confidence affects individual self-evolution updates. Figure~\ref{fig:qualitative} compares a high-confidence correct rollout with a low-confidence erroneous one. The important point is not that COSE can perfectly identify every wrong judgment; it cannot. Rather, the example shows how COSE changes the consequence of an uncertain judgment. In a standard self-judging loop, an incorrect endorsement can produce a full-strength positive update. In COSE, the same endorsement receives a smaller Solver weight when the Judge is uncertain, so the model is less likely to reinforce the faulty reasoning pattern. This qualitative behavior matches the ablation results: the largest gains come from controlling gradient strength, not simply from filtering examples or generating more tasks.

\section{Conclusion}
\label{sec:conclusion}

We presented COSE, a confidence-orchestrated framework for self-evolving reasoning training. COSE addresses noisy LLM-generated validation and judging feedback by estimating confidence from Validator and Judge outputs and using it to control learning through confidence-weighted PPO and confidence-prioritized replay. Across 19 held-out benchmarks and four Qwen/Llama backbones, COSE improves over base models and remains competitive with or stronger than prior self-evolving baselines. Ablations show that confidence-weighted PPO is the key contributor, while replay priority provides additional gains. Overall, COSE shows that self-evolution becomes more effective when models not only learn from their own feedback, but also control how strongly uncertain feedback affects training.

\section*{Limitations}
\label{sec:limitations}

COSE reduces the influence of uncertain LLM-generated feedback, but it does not guarantee that the feedback is correct. High-confidence Validator or Judge outputs can still be wrong, especially on adversarial, ambiguous, or out-of-distribution reasoning problems. Confidence should therefore be understood as a heuristic uncertainty signal rather than a formal correctness certificate. This limitation is particularly important in self-evolution, where repeated training on confidently wrong feedback may still reinforce errors over multiple rounds.

COSE also relies on access to token-level probability distributions. This makes the method well suited to open-weight models and inference setups that expose logits, but less directly applicable to black-box APIs that only return generated text. While alternative confidence signals such as self-consistency or verbalized confidence could be used in black-box settings, they may require additional sampling and may not provide the same fine-grained signal as token-level entropy.

Another limitation is that our experiments focus on relatively small open-weight backbones, ranging from 0.6B to 4B parameters. These models are important because self-evolution is especially useful when strong human-curated supervision is unavailable, but our results do not fully characterize scaling behavior. Larger models may produce less noisy validation and judging feedback, which could reduce the marginal benefit of confidence weighting, or they may generate more persuasive but still incorrect judgments, which could make confidence control even more important. Evaluating COSE on larger models and longer self-evolution runs is an important direction for future work.

Our current implementation also uses a simple confidence computation based on normalized entropy and sequence-level aggregation. Although the hyperparameter study suggests that COSE is not highly sensitive to the exact confidence estimator, more calibrated confidence measures may further improve training stability. In particular, future work could combine token-level confidence with semantic uncertainty, cross-sample agreement, or external verifiers when available.

Finally, COSE is designed for reasoning tasks and is evaluated on general reasoning, mathematics, and code generation benchmarks. It is not intended to replace domain-specific verification in high-stakes settings. For applications such as medicine, law, finance, or safety-critical engineering, confidence-weighted self-evolution should be combined with expert review, external tools, and task-specific validation protocols.

\section*{Ethical Considerations}
\label{sec:ethics}

COSE studies how models can improve from self-generated tasks and feedback. This can reduce dependence on costly human annotation, but it also raises risks if self-evolved models amplify incorrect reasoning, biased judgments, or harmful problem distributions. Our experiments use standard reasoning, mathematics, and code benchmarks and do not involve private user data or human-subject annotation. Nevertheless, COSE should be deployed cautiously: confidence-weighted feedback is not a substitute for human oversight or external verification in high-stakes domains. We encourage future work to audit self-generated training data, monitor bias and failure modes across self-evolution rounds, and clearly document when model-generated feedback is used as supervision.
\bibliography{custom}

\newpage
\appendix

This appendix provides additional details and results that support the main paper. 
Appendix~\ref{app:related} gives a detailed discussion of related work, including self-evolving LLMs, LLM-as-Judge behavior, confidence estimation, and sample weighting. 
Appendix~\ref{app:preliminaries} provides background on SFT, RLVR, and self-evolution. 
Appendix~\ref{app:exp_setup} reports experimental setup details, including benchmarks, evaluation settings, and training hyperparameters. 
Appendix~\ref{app:math} provides the full mathematics results across all four backbones. 
Appendix~\ref{app:code} provides the full code results. 
Appendix~\ref{app:dynamics} presents per-benchmark training dynamics. 
Appendix~\ref{app:prompts} includes the role prompts used by the Proposer, Validator, Solver, and Judge. 
Appendix~\ref{app:eval_prompts} describes evaluation prompts. 
Appendix~\ref{app:notation} summarizes the notation used by COSE. 
Appendix~\ref{app:artifacts} documents artifact and license information. 
Appendix~\ref{app:compute} reports compute infrastructure and API-based evaluation cost.

\section{Detailed Related Work}
\label{app:related}

\paragraph{Self-evolving LLMs.}
Self-evolving frameworks train a model to generate, attempt, and learn from its own tasks, reducing the need for human-curated supervision. Externally grounded methods obtain feedback from verifiers or environments, such as Python execution in AZR~\citep{zhao2025absolute} or retrieval over a document corpus in SPICE~\citep{liu2025spice}. These signals are reliable within their supported domains, but they do not generalize to reasoning tasks without executable or retrieval-grounded verification. Self-judging methods such as MAE~\citep{chen2025mae} and R-Zero~\citep{huang2025rzero} broaden the scope by using model-generated feedback, but then inherit the uncertain feedback limits of LLM judges. Other work improves self-evolution with role specialization, environment grounding, diversity preservation, and learnability mechanisms~\citep{chen2025spc,liu2025toolr0,wang2025cure,li2025rdiverse,liu2026learnable}. COSE is complementary to these directions: it focuses on how much self-generated validation and judging feedback should be trusted during learning.

\paragraph{LLMs judging their own outputs.}
LLM-as-Judge has become a common evaluation paradigm~\citep{zheng2023judging}, but many studies show that LLM judgments are biased and noisy. Models can prefer their own outputs~\citep{panickssery2024selfpreference,xu2024perils}, depend on answer position or style~\citep{wang2024positionbias}, and struggle most on harder reasoning tasks~\citep{stechly2024selfverification,huang2024cannotcorrect}. Surveys further document the strengths and limitations of LLM-based evaluation~\citep{gu2024llmasjudge,li2024llmevaluator}. These limitations are especially important for self-evolution because validation and judging errors are not merely evaluation noise; they become reward signals that can be amplified by later policy updates.

\paragraph{Confidence estimation for model feedback.}
Confidence estimation has long been studied through predictive distributions. Maximum softmax probability and logit-based scores are standard baselines for identifying misclassified or out-of-distribution examples~\citep{hendrycks2017baseline}, while calibration work shows that neural probabilities can be overconfident and require correction~\citep{guo2017calibration}. Entropy-based methods use predictive entropy or normalized entropy to measure uncertainty in the output distribution~\citep{geifman2017selective,malinin2018predictive}. For LLMs, recent work studies confidence through token probabilities, semantic uncertainty, and consistency across samples~\citep{kadavath2022languagemodelsmostlyknow,kuhn2023semantic,lin2024generating,manakul2023selfcheckgpt,geng2024survey}. COSE adopts this broad view of confidence, but uses it as a training-control signal: Validator and Judge confidence determines how strongly self-generated feedback affects PPO updates and replay sampling.

\paragraph{Sample weighting and prioritized replay.}
COSE uses confidence through sample weighting and prioritized replay, both of which are standard optimization tools. Prioritized experience replay increases the sampling probability of more useful transitions, often using TD error as the priority signal~\citep{schaul2016per,horgan2018apex}. PPO uses importance ratios to stabilize policy optimization~\citep{schulman2017ppo}, and preference-optimization methods weight updates by reward-model confidence or preference strength~\citep{coste2024rewardensembles,rafailov2023dpo,wu2024beta}. COSE differs in the source of the weighting signal. Instead of relying on TD error or external reward confidence, it uses intrinsic confidence from the same Validator and Judge that produce self-evolution feedback, making the learning loop aware of when its own feedback is uncertain.

\section{Preliminaries}
\label{app:preliminaries}

We situate COSE against two training paradigms it builds on.
\emph{Supervised fine-tuning} (SFT) minimizes the conditional
negative log-likelihood
$\mathcal{L}_{\text{SFT}}(\theta) = -\mathbb{E}_{(x,c^\star,y^\star)\sim\mathcal{D}}
\log\pi_\theta(c^\star,y^\star\mid x)$
on human-annotated demonstrations $(x,c^\star,y^\star)$ of queries,
chain-of-thought rationales, and reference answers.
\emph{Reinforcement learning with verifiable rewards} (RLVR) relaxes
this to $(x,y^\star)$ pairs by having the model generate its own
traces and rewarding outputs against an external verifier $r(y,y^\star)$,
\begin{equation}
\mathcal{J}_{\text{RLVR}}(\theta) =
\mathbb{E}_{(x,y^\star)\sim\mathcal{D},\,y\sim\pi_\theta(\cdot\mid x)}
\bigl[r(y,y^\star)\bigr].
\end{equation}

Self-evolving methods~\citep{zhao2025absolute,huang2025rzero,liu2025spice}
go one step further, removing the human-curated $(x,y^\star)$ entirely
and replacing $r$ with a model-based estimate $\hat{r}(y)=J_\theta(y)$,
where the Judge $J_\theta$ is the same policy under an evaluation
prompt:
\begin{equation}
\mathcal{J}_{\text{self}}(\theta) =
\mathbb{E}_{x\sim\mathcal{D},\,y\sim\pi_\theta(\cdot\mid x)}
\bigl[J_\theta(y)\bigr].
\end{equation}
The price of this autonomy is that every Judge verdict becomes a
gradient update, and the generator and Judge share
parameters~\citep{stechly2024selfverification,huang2024cannotcorrect}:
the errors the Judge fails to detect are precisely the errors training
will reinforce. The signal needed to identify those errors---per-sample
token-level confidence---is already produced at every rollout but
discarded by existing self-evolving frameworks.

\section{Experimental Setup}
\label{app:exp_setup}
\paragraph{Evaluation benchmarks.} We evaluate on 19 held-out benchmarks organized by domain.
\emph{General reasoning} comprises 10 benchmarks: MMLU~\citep{hendrycks2021mmlu}, GPQA~\citep{rein2024gpqa}, BBH~\citep{suzgun2023bbh}, LiveBench Reasoning~\citep{white2024livebench}, ARC-Challenge~\citep{clark2018arc}, CommonsenseQA~\citep{talmor2019commonsenseqa}, OpenBookQA~\citep{mihaylov2018openbookqa}, Natural Questions~\citep{kwiatkowski2019nq}, TriviaQA~\citep{joshi2017triviaqa}, and SQuAD~\citep{rajpurkar2016squad}.
\emph{Mathematics} comprises 5 benchmarks: GSM8K~\citep{cobbe2021gsm8k}, MATH~\citep{hendrycks2021math}, AMC, Minerva~\citep{lewkowycz2022minerva}, and Olympiad-Bench~\citep{he2024olympiad}.
\emph{Code} comprises 4 benchmarks: HumanEval and HumanEval+, MBPP and MBPP+, all from the EvalPlus suite~\citep{liu2023evalplus}.
None of these benchmarks appear in the COSE training loop; all evaluation is on held-out data.

\paragraph{Evaluation protocol.}
We use greedy decoding ($T{=}0$, $n{=}1$) for all evaluations. Code benchmarks are evaluated with EvalPlus and reported as pass@1 percentages. For mathematics and general reasoning benchmarks, we follow the evaluation protocol of MAE~\citep{chen2025mae} and use \texttt{gpt-4.1-nano} as the judge with a fixed scoring template. The judge receives the problem, the model prediction, and the reference answer when available, and determines whether the prediction should be counted as correct. We report these results as GPT-judge accuracy. We use the same evaluation prompts, decoding settings, judge model, and scoring procedure for all methods. Self-generated training questions are not sampled from evaluation benchmarks, and we apply benchmark-name and exact-match filtering against held-out evaluation data to reduce contamination risk.

\paragraph{Hyperparameters.} We train each base model for a fixed step budget---260 steps for Qwen3-0.6B, 200 steps for Qwen3-4B, 100 steps for Llama-3.2-3B-Instruct, and 100 steps for Qwen2.5-3B-Instruct---selected to match the compute used by the baselines. PPO uses clip ratio $\epsilon{=}0.2$, learning rate ${1{\times}10^{-6}}$ with cosine decay, KL coefficient ${0.01}$ against the reference policy, batch size~${128}$, and a per-sample weight floor of $0.1$ (Eq.~\ref{eq:weights}). The replay buffer~$\mathcal{B}$ has capacity~${8\text{K}}$ and is sampled proportionally to Eq.~\ref{eq:priority}. The Proposer/Validator phase runs every $K{=}1$ step.

\section{Math Results}
\label{app:math}

Table~\ref{tab:math_results} in the main body reports mathematics
performance on Llama-3.2-3B and Qwen3-4B; the full table covering all
four backbones is reproduced as Table~\ref{app:math_results} here. We use
the expanded view to make three observations that the main body does not
have space for and that tighten the argument of
Section~\ref{sec:rq2_math}.

\paragraph{The size of COSE's advantage tracks the difficulty of the
benchmark, not the scale of the backbone.}
On Qwen3-0.6B, COSE's gains over the strongest baseline grow with
benchmark difficulty: $+7.4$ on GSM8K, $+13.8$ on MATH, $+15.85$ on AMC,
$+22.78$ on Minerva, and $+38.18$ on Olympiad-Bench. The same pattern
holds on Llama-3.2-3B ($+13.8$, $+27.2$, $+29.28$, $+18.96$, $+24.40$),
and a muted version on Qwen3-4B. By contrast, AZR, MAE, and R-Zero
improve roughly uniformly across difficulty levels. This is the
diagnostic for our central claim. On easy benchmarks (GSM8K), a model can
solve most problems correctly and the Judge can verify most of those
solutions, so any reasonable self-evolution method recovers most of the
available signal. On hard benchmarks (Olympiad-Bench), correct solutions
are rare, incorrect-but-plausible solutions are common, and the Judge's
endorsement-versus-rejection decisions matter much more. A method that
treats all accepted judgments as equally trustworthy will absorb both the
correct and the confidently-wrong endorsements; a method that scales
gradient by Judge confidence absorbs proportionally less of the wrong
ones. The widening gap on harder benchmarks is therefore not a separate
phenomenon---it is the per-problem confidence weighting compounding over
many updates.

\paragraph{Confidence weighting helps most precisely where it should.}
The argument above predicts that the Qwen2.5-3B-Instruct results, where
COSE only matches the strongest baseline ($+1.5$ avg.\ over MAE), are not
a counterexample but a confirmation. Qwen2.5-3B-Instruct starts higher on
mathematics than Llama-3.2-3B (49.52 vs.\ 47.98 avg.) but evolves less
under every method, including the baselines: AZR gains 2.78, MAE gains
3.60, COSE gains 5.12. This consistent ceiling across methods indicates
that the bottleneck for Qwen2.5-3B-Instruct is the model's reasoning
capability rather than Judge uncertainty, so improving the latter
produces correspondingly small returns. The single instance where COSE
loses to a baseline on this backbone---AMC, where MAE scores 41.80 vs.\
COSE's 39.17---is consistent with this reading: AMC has the smallest
absolute headroom of any benchmark on this backbone (Base 39.76, ceiling
above), so a difference of 2.6 points is within the band where any
method's update strategy can win or lose on noise. The corresponding gain
of $+12.96$ on Minerva for COSE on the same backbone tells the more
important story.

\paragraph{Scale does not eliminate the value of confidence control on
hard math.}
On Qwen3-4B, the absolute gains are smaller because the backbone starts
near saturation ($74.82$ avg.\ for Base, vs.\ ceilings around 95--100),
but their distribution remains diagnostic. The benchmarks with the most
remaining headroom---Olympiad-Bench (Base $66.80$) and Minerva
(Base $78.70$)---show the largest COSE gains over the next-best baseline
($+9.32$ and $+6.58$, respectively), while the more saturated GSM8K and
MATH show $+1.20$ and $+3.20$. We read this as evidence that confidence
weighting does not provide a fixed multiplicative improvement; it
provides a noise-floor reduction that becomes irrelevant when the noise
is already small. For Olympiad-Bench, where even a 4B model fails on
most problems and the Judge faces hard endorsement decisions on every
attempt, the floor reduction is large; for GSM8K, where solutions are
shorter and the Judge is more often correct on its own attempts, the
reduction is small. This is the same mechanism that drives the
Qwen3-0.6B difficulty-scaling pattern, observed in a different regime.

\paragraph{Implications.} The math table thus encodes a stronger claim
than ``COSE is best on average.'' It encodes a regularity: \emph{the
benefit of training-control over feedback grows with how noisy that
feedback is}. Difficulty proxies for uncertainty when the Judge is the
same model as the Solver, scale proxies for it in the opposite direction,
and benchmark headroom interacts with both. The four-backbone view makes
this regularity visible in a way that any single backbone could not.

\begin{table}[t]
\centering
\small
\setlength{\tabcolsep}{1pt}
\begin{tabular}{l|ccccc|c}
\toprule
\rowcolor{t3head}
\textbf{Method} & \textbf{GSM8K} & \textbf{MATH} & \textbf{AMC} & \textbf{Minerva} & \textbf{Olympiad} & \textbf{Avg.} \\
\midrule
\rowcolor{t3group} \multicolumn{7}{c}{\textit{Qwen3-0.6B}} \\
\midrule
Base          & 81.00 & 67.00 & 41.00 & 48.80 & 30.40 & 53.64 \\
AZR           & 84.20 & 70.40 & 45.30 & 53.60 & 36.80 & 58.06 \\
MAE           & 85.60 & 72.20 & 47.80 & 55.40 & 39.20 & 60.04 \\
R-Zero        & 84.80 & 71.60 & 46.90 & 54.80 & 38.40 & 59.30 \\
\rowcolor{t3cose} \textbf{COSE} & \textbf{93.00} & \textbf{86.00} & \textbf{63.65} & \textbf{78.18} & \textbf{77.38} & \textbf{79.64} \\
\midrule
\rowcolor{t3group} \multicolumn{7}{c}{\textit{Qwen2.5-3B-Instruct}} \\
\midrule
Base          & 85.20 & 60.40 & 39.76 & 34.52 & 27.73 & 49.52 \\
AZR           & 86.40 & 62.40 & 41.10 & 41.22 & 30.40 & 52.30 \\
MAE           & 86.80 & 63.20 & \textbf{41.80} & 42.60 & 31.20 & 53.12 \\
R-Zero        & 86.00 & 62.80 & 41.30 & 40.90 & 30.80 & 52.36 \\
\rowcolor{t3cose} \textbf{COSE} & \textbf{87.33} & \textbf{67.67} & 39.17 & \textbf{47.02} & \textbf{32.00} & \textbf{54.64} \\
\midrule
\rowcolor{t3group} \multicolumn{7}{c}{\textit{Llama-3.2-3B-Instruct}} \\
\midrule
Base          & 63.00 & 47.00 & 27.70 & 64.10 & 38.10 & 47.98 \\
AZR           & 72.40 & 56.20 & 36.80 & 68.40 & 44.20 & 55.60 \\
MAE           & 75.20 & 59.80 & 40.60 & 70.20 & 47.60 & 58.68 \\
R-Zero        & 73.80 & 58.40 & 38.90 & 69.10 & 46.30 & 57.30 \\
\rowcolor{t3cose} \textbf{COSE} & \textbf{89.00} & \textbf{87.00} & \textbf{69.88} & \textbf{89.16} & \textbf{72.00} & \textbf{81.41} \\
\midrule
\rowcolor{t3group} \multicolumn{7}{c}{\textit{Qwen3-4B}} \\
\midrule
Base          & 85.00 & 81.00 & 62.60 & 78.70 & 66.80 & 74.82 \\
AZR           & 93.20 & 82.40 & 64.80 & 81.10 & 68.60 & 78.02 \\
MAE           & 92.80 & 83.20 & 65.60 & 81.80 & 69.40 & 78.56 \\
R-Zero        & 92.40 & 82.60 & 64.90 & 81.20 & 68.90 & 78.00 \\
\rowcolor{t3cose} \textbf{COSE} & \textbf{94.40} & \textbf{86.40} & \textbf{68.80} & \textbf{88.38} & \textbf{78.72} & \textbf{83.34} \\
\bottomrule
\end{tabular}
\caption{Math results across four base models.}
\label{app:math_results}
\vspace{-10pt}
\end{table}

\section{Code Results}
\label{app:code}

Code is the domain where COSE has the competitive theoretical case: HumanEval,
HumanEval+, MBPP, and MBPP+ all admit deterministic correctness via test
execution, which is the kind of feedback COSE's confidence weighting is
designed to substitute for. We include the full four-backbone code table
(Table~\ref{app:code_full}) here not because it shows large gains but
because it tests a specific failure mode that any training-control method
must avoid: degrading performance in regimes where its mechanism does not
apply.

\paragraph{No backbone regresses meaningfully under COSE.}
The largest negative delta from base in the entire table is $-1.18$
(Llama-3.2-3B HumanEval), and the largest negative delta from any
baseline is $-2.20$ (Llama-3.2-3B HumanEval vs.\ AZR). Every other COSE
cell is within $\pm 1.5$ of base. When feedback is already reliable, scaling gradient by Judge confidence
reduces to a near-uniform update rule, so the optimization trajectory
should be roughly the same as standard PPO. Code confirms this---COSE
neither helps nor harms substantially on the saturated benchmarks (MBPP,
Qwen3-4B HumanEval), and tracks the noise band of the baselines on the
rest. A method that achieved its math and reasoning gains by overfitting
to a domain-specific update rule would show systematic code degradation;
COSE does not.

\paragraph{Where COSE leads, it leads on the weakest backbones.}
The two backbones where COSE wins on average---Qwen3-0.6B
($+1.65$ over R-Zero) and Qwen2.5-3B-Instruct ($+0.15$ over R-Zero)---are
exactly the backbones whose starting code accuracy is lowest (44.12 and
69.47, vs.\ 54.02 and 74.80 for the others). This is the same
difficulty-scaling pattern we identify for mathematics in
Appendix~\ref{app:math}, expressed here in a much narrower band. A 0.6B
model writing code still makes errors that an in-model Judge can usefully
flag, even with execution available downstream; on stronger backbones,
the Judge's marginal contribution shrinks because the Solver is already
mostly correct. The gains on Qwen3-0.6B are modest in absolute terms
($+4.21$ avg.\ over base) but they confirm that confidence weighting and
test execution are not antagonistic: when there is residual noise for
trust control to act on, COSE finds it.

\paragraph{AZR's advantage on Llama-3.2-3B is consistent with the design.}
AZR is the strongest method on Llama-3.2-3B code ($+2.38$ over base on
average), and we read this as a feature of the comparison, not a
counterexample. AZR uses Python execution as its training signal; code is
the regime where this signal is most direct, and we should expect a
verifier-grounded method to be competitive or stronger here. The
important observation is that COSE does not lose substantially in this
regime---0.55 above base, 1.85 below AZR---despite being designed for a
fundamentally different feedback regime. This is the boundary we draw in
the main body: COSE is most valuable when verification is weak, but
remains compatible with domains where stronger feedback exists.
\begin{table}[t]
\centering
\setlength{\tabcolsep}{2pt}
\small
\begin{tabular}{lccccc}
\toprule
\rowcolor{t2head}
Method & HumanEval & HumanEval+ & MBPP & MBPP+ & Avg. \\
\midrule
\rowcolor{t2group} \multicolumn{6}{c}{\textit{Qwen3-4B}} \\
\midrule
Base   & 78.66 & 73.17 & 79.63 & 67.72 & 74.80 \\
AZR    & \textbf{80.50} & \textbf{75.00} & 79.90 & 67.50 & \textbf{75.73} \\
R-Zero & 78.00 & 73.20 & \textbf{80.40} & \textbf{68.30} & 74.98 \\
\rowcolor{t2cose} COSE   & 79.30 & 73.80 & 79.90 & 67.50 & 75.13 \\
\midrule
\rowcolor{t2group} \multicolumn{6}{c}{\textit{Qwen3-0.6B}} \\
\midrule
Base   & 40.24 & 35.98 & 54.50 & 45.77 & 44.12 \\
AZR    & 39.60 & 36.60 & 54.50 & 46.80 & 44.38 \\
R-Zero & 42.70 & 39.00 & \textbf{56.30} & 48.70 & 46.68 \\
\rowcolor{t2cose} COSE   & \textbf{45.70} & \textbf{41.50} & 56.10 & \textbf{50.00} & \textbf{48.33} \\
\midrule
\rowcolor{t2group} \multicolumn{6}{c}{\textit{Llama-3.2-3B-Instruct}} \\
\midrule
Base   & 54.88 & 50.61 & 61.11 & 49.47 & 54.02 \\
AZR    & \textbf{55.50} & \textbf{52.40} & \textbf{63.20} & \textbf{54.50} & \textbf{56.40} \\
R-Zero & 54.90 & 50.60 & 60.60 & 50.50 & 54.15 \\
\rowcolor{t2cose} COSE   & 53.70 & 49.40 & \textbf{63.20} & 51.90 & 54.55 \\
\midrule
\rowcolor{t2group} \multicolumn{6}{c}{\textit{Qwen2.5-3B-Instruct}} \\
\midrule
Base   & 73.78 & 68.90 & 73.28 & 61.90 & 69.47 \\
AZR    & 72.60 & 68.90 & 73.50 & 61.90 & 69.23 \\
R-Zero & 74.40 & \textbf{70.70} & \textbf{74.60} & 61.90 & 70.40 \\
\rowcolor{t2cose} COSE   & \textbf{75.00} & \textbf{70.70} & 73.50 & \textbf{63.00} & \textbf{70.55} \\
\bottomrule
\end{tabular}
\caption{Code benchmark results across four backbones. }
\label{app:code_full}
\vspace{-10pt}
\end{table}

\section{Per-Benchmark Training Dynamics}
\label{app:dynamics}

Figure~\ref{fig:training_curves} in the main body aggregates COSE training
into three domain averages per backbone. Aggregation is useful for headline
claims but hides exactly the failure modes that motivate confidence
weighting in the first place: averages smooth over benchmarks where gains
and losses cancel, and they suppress the within-run instabilities that
would dominate a single-benchmark view. This appendix presents the
per-benchmark training curves, one figure per backbone. Each panel shows
held-out accuracy for one benchmark with an auto-zoomed y-axis (so small
movements remain legible), a dashed reference at the step-0 starting
accuracy, and a corner annotation reporting final$-$initial $\Delta$
(green for $\Delta>0.005$, red for $\Delta<-0.005$, grey for ties).
Top-spine color encodes the benchmark domain. Benchmark order is identical
across the three figures, so panels at the same grid position correspond
directly between backbones.

We read these three figures as three different operating regimes for the
training-control argument of Section~\ref{sec:method}: a regime where naive
self-evolution would fail and COSE recovers (Qwen3-0.6B), a regime where
self-evolution starts well but distribution shift inside the replay loop
re-introduces noise (Llama-3.2-3B), and a regime where COSE has limited
room to act because the backbone is near saturation (Qwen3-4B).

\subsection{Qwen3-0.6B: confidence weighting as an enabling condition,
not a refinement}
\label{app:dyn_qwen06}

\begin{figure*}[t]
  \centering
  \includegraphics[width=0.98\linewidth]{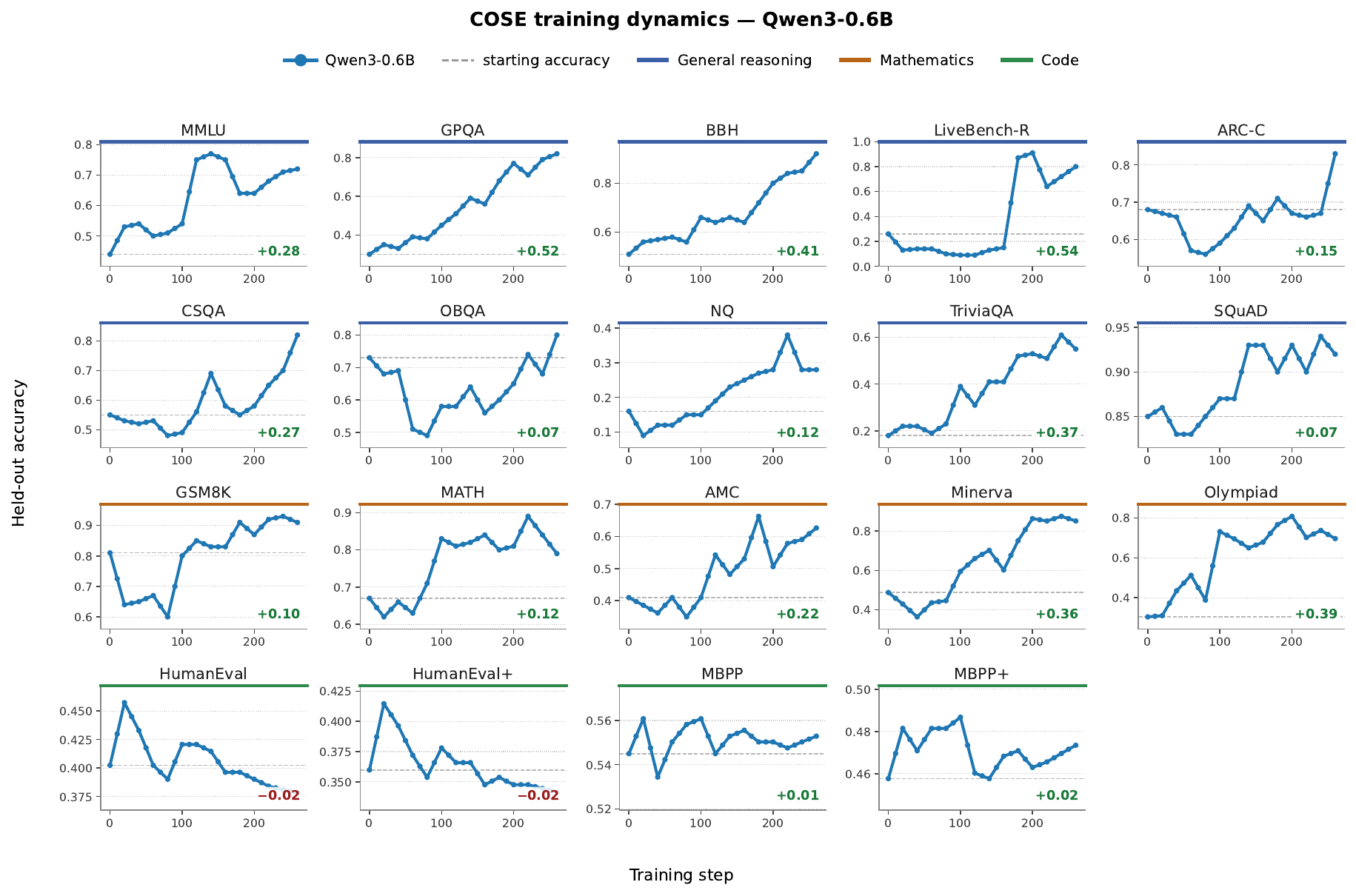}
  \caption{Per-benchmark training dynamics under COSE on Qwen3-0.6B.}
  \label{fig:dynamics_qwen06}
\end{figure*}

The Qwen3-0.6B figure (Figure~\ref{fig:dynamics_qwen06}) is the strongest
evidence that confidence weighting changes the qualitative behavior of
self-evolution rather than merely smoothing it. The model starts at 0.47
overall and ends at 0.74; this would be unremarkable if the per-benchmark
view were monotonic, but the panels reveal a more interesting pattern.

Several benchmarks (MMLU, OBQA, CSQA, ARC-C, GSM8K) decline in the first
40--80 steps before recovering and improving. A standard self-evolving
method would interpret this early decline as a stable failure mode and
keep reinforcing it: the Judge endorses incorrect early answers, those
endorsements become full-strength positive gradients, and the model
locks into the wrong region of policy space. The ablation in
Table~\ref{tab:ablation} supports this reading---removing confidence
weighting costs over 10 absolute points on math for this backbone, far
more than removing replay priority. What the per-benchmark view adds is a
mechanism: the early-dip-then-recovery pattern is exactly what
confidence-weighted PPO is designed to produce. Low-confidence endorsements
during the noisy early phase have their gradients dampened to the
$w=0.1$ floor, the model avoids consolidating around those endorsements,
and the replay buffer is given time to accumulate a more reliably validated
question pool. Once Validator confidence on accepted questions rises, the
intermediate-difficulty term in Eq.~\ref{eq:priority} takes over and the
Solver begins climbing.

The LiveBench-R panel is the clearest case. The curve sits near 0.15 for
roughly 150 steps---essentially no learning signal---and then jumps to
$\sim$0.9. The horizontal phase is not a failure of COSE; it is a
correctly damped phase. The Solver cannot yet reason at the depth
LiveBench-R requires, the Judge is uncertain on its own attempts, and
confidence weighting prevents that uncertainty from being baked into the
policy. When the model finally acquires the relevant reasoning patterns
from other benchmarks, LiveBench-R unlocks discontinuously rather than
having to recover from accumulated bad updates.

The four code panels move by less than 0.08 each. We read this as the
control case: when feedback is essentially reliable (HumanEval and MBPP
have closed-form correctness), confidence weighting reduces to a
near-uniform update rule and the method neither helps nor hurts. The flat
code curves therefore support the broader claim of
Section~\ref{sec:rq2_code}: COSE does not pay for its gains on noisy
domains with regressions on clean ones.

\subsection{Llama-3.2-3B: confidence weighting is a brake, not a cure}
\label{app:dyn_llama}

\begin{figure*}[t]
  \centering
  \includegraphics[width=0.98\linewidth]{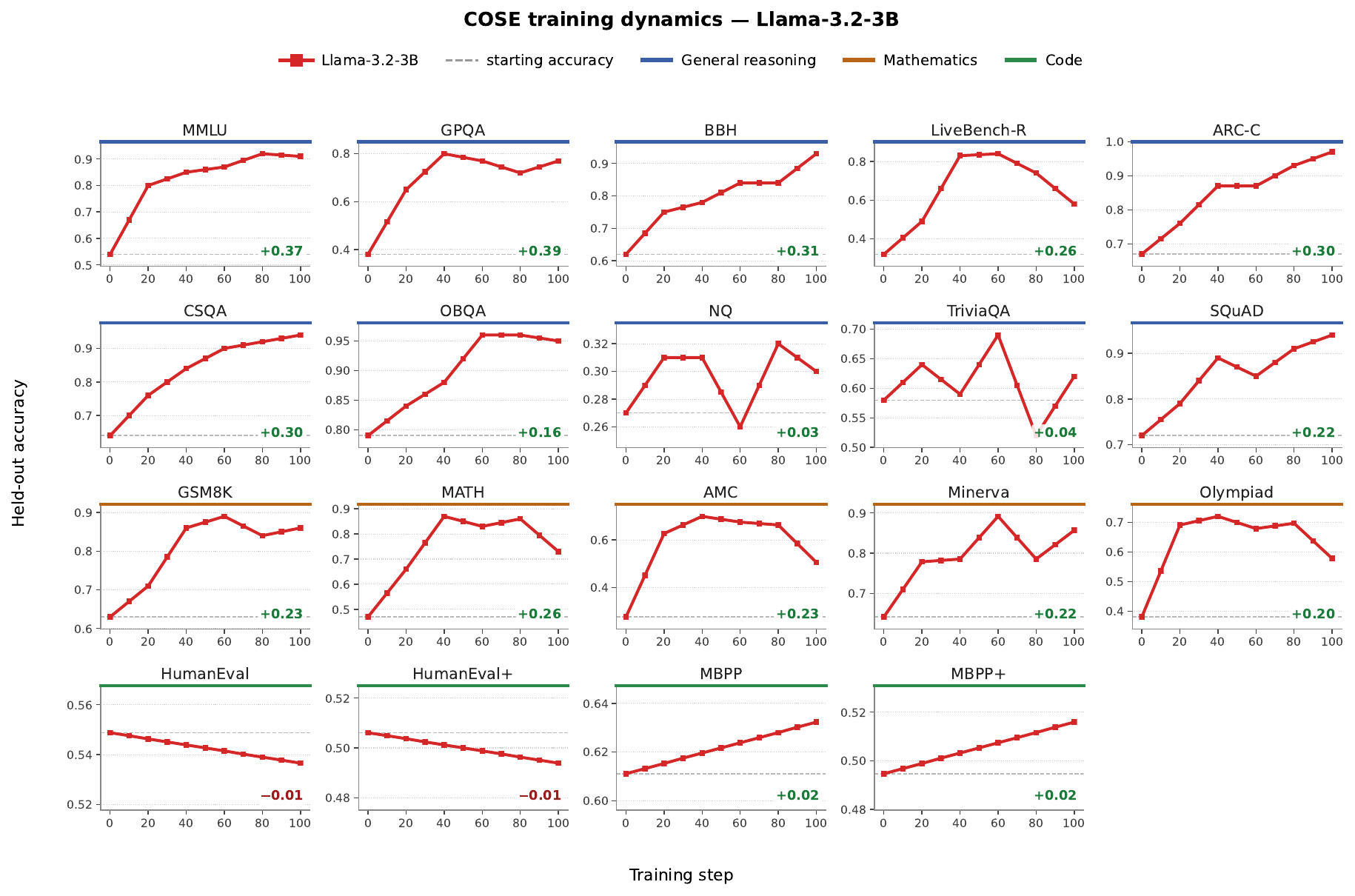}
  \caption{Per-benchmark training dynamics under COSE on Llama-3.2-3B.
  Panel layout, annotation, and color encoding match
  \Cref{fig:dynamics_qwen06}.}
  \label{fig:dynamics_llama3b}
\end{figure*}

The Llama-3.2-3B figure (Figure~\ref{fig:dynamics_llama3b}) makes a less
favorable but more important point: confidence weighting bounds the damage
of noisy feedback but does not eliminate it. Five benchmarks
(LiveBench-R, MATH, AMC, Minerva, Olympiad) climb sharply in the first
40--60 steps, peak, and then partially decline. NQ and TriviaQA oscillate
without a clean trend.

The shape of these declines is the diagnostic. They are gradual rather
than catastrophic---typically 0.05--0.15 absolute over 40 steps---and they
concentrate in mathematics, the domain where Judge feedback is most
fragile because solution-level correctness can hinge on subtle
intermediate errors that a model-as-Judge fails to detect. This matches
the mechanism we argue for in Section~\ref{sec:rq2_math}: math is the
high-risk regime not because math is hard, but because plausible-looking
incorrect derivations receive confidently wrong endorsements. Confidence
weighting reduces the magnitude of each individual damaging update, which
slows the decline, but it cannot correct a Judge that is systematically
miscalibrated on a class of derivations. Over enough replays of similar
question types, even down-weighted bad gradients accumulate.

\subsection{Qwen3-4B: a saturation test for the training-control claim}
\label{app:dyn_qwen4}

\begin{figure*}[t]
  \centering
  \includegraphics[width=0.98\linewidth]{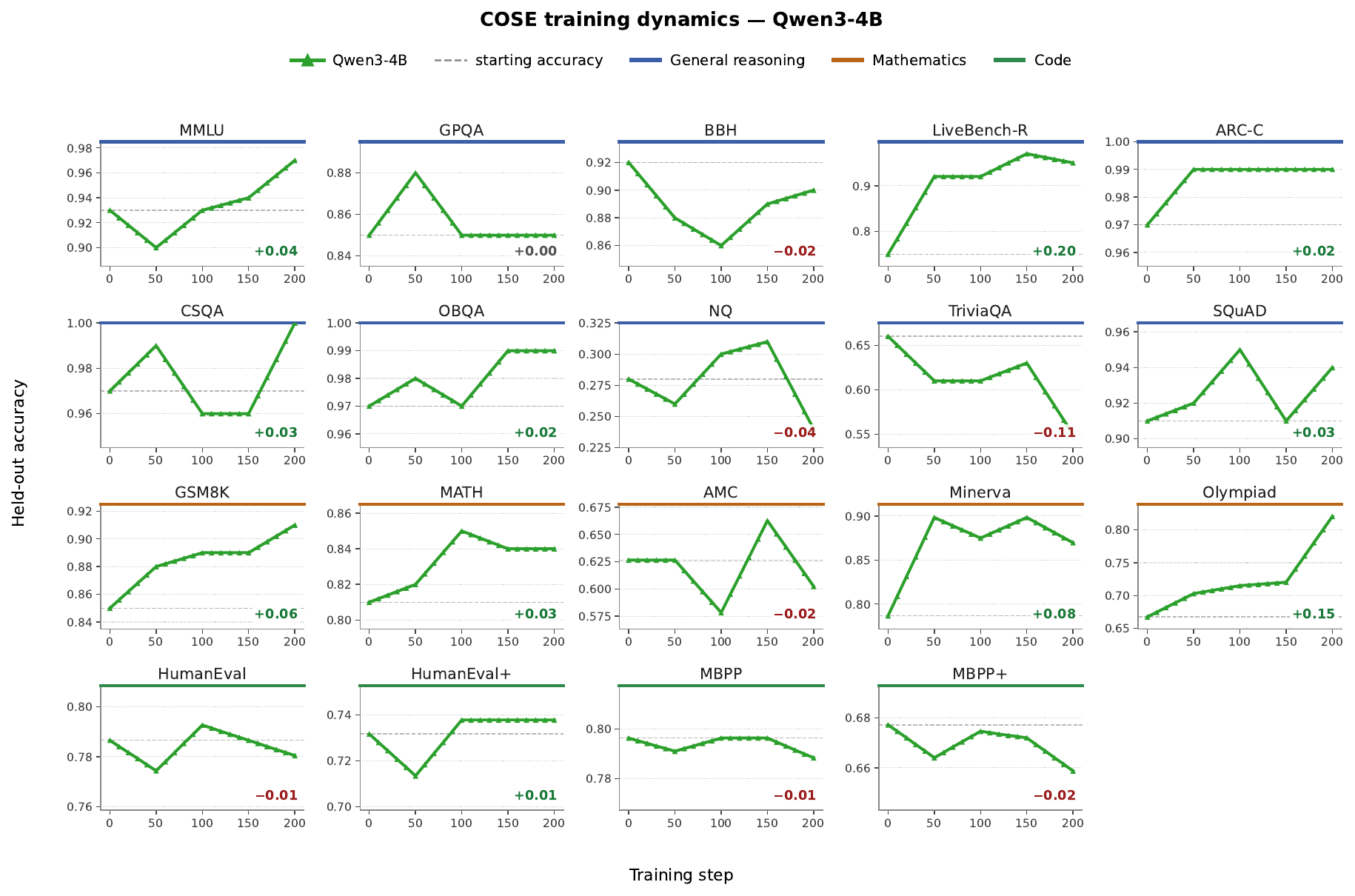}
  \caption{Per-benchmark training dynamics under COSE on Qwen3-4B.
  Layout matches \Cref{fig:dynamics_qwen06}. }
  \label{fig:dynamics_qwen4b}
\end{figure*}

Qwen3-4B (Figure~\ref{fig:dynamics_qwen4b}) is the regime where COSE's
gains \emph{should} be smallest, and they are. Eleven of nineteen panels
move by less than 0.05 absolute. The interest in this figure is not in the
deltas but in what the regime tells us about the method.

A method that strongly reinforces accepted feedback would push a
near-saturated model into systematic regression: the Validator and Judge,
operating near their own ceiling, would generate confidently wrong
endorsements on borderline cases, and a method without trust control would
treat those endorsements as ground truth. COSE does not. The largest
gains appear exactly where headroom is largest---LiveBench-R from 0.75 to
0.95, Olympiad from 0.67 to 0.82---while the saturated benchmarks
fluctuate within narrow bands rather than degrading. GPQA in particular
ends at exactly its starting value of 0.85 after touching 0.88
mid-training, which is the cleanest possible signal that COSE will not
destabilize a benchmark just because training continues.

The one substantive regression---TriviaQA, $-0.11$---deserves a separate
explanation, because it is not predicted by the training-control argument
and is the largest negative delta in any of our experiments. We attribute
it to curriculum drift rather than to feedback uncertainty. The Proposer
prompt (Appendix~\ref{app:prompts}) explicitly discourages trivia-style
and web-dependent questions, so the self-generated training distribution
moves away from the regime TriviaQA samples. Compounding this, training
on multi-step reasoning appears to shift the model's output style toward
longer, more deliberative answers, which can hurt short factual recall
where a concise answer is preferred. NQ shows a smaller version of the
same effect ($-0.04$). Neither regression is a failure of confidence
weighting per se---a confidence-aware method has no reason to detect or
correct a mismatch between the curriculum and the eval distribution---but
they highlight a real limitation that we discuss in
Section~\ref{sec:limitations}: COSE controls how aggressively the model
trusts its own feedback, not what its own feedback is about.

\subsection{What the per-benchmark view adds to the paper's argument}
\label{app:dyn_summary}

Read together, the three figures clarify what confidence-weighted
self-evolution does and does not do. It changes how the optimization
trajectory responds to noisy feedback (Qwen3-0.6B: noisy early phases
are damped rather than baked in); it bounds the rate at which systematic
Judge miscalibration corrupts a policy (Llama-3.2-3B: math curves decline
gradually rather than catastrophically); and it stays out of the way when
feedback is already reliable or when the model has little headroom
(Qwen3-4B and the code rows on all three backbones). It does \emph{not}
correct mismatches between the self-generated curriculum and the held-out
evaluation distribution (Qwen3-4B TriviaQA), and it does not guarantee
that the final-step checkpoint is the best checkpoint (Llama-3.2-3B math).
These are exactly the boundary conditions stated in
Section~\ref{sec:limitations}, now visible in the curves.

\section{Role Prompts}
\label{app:prompts}

This appendix summarizes the prompt templates used for the four roles in COSE. To avoid formatting overflow in the two-column EMNLP layout, we present abbreviated versions capturing the essential structure and constraints.
In our prompts, the Validator and Judge output integer scores on a fixed scale; we normalize these scores to $[0,1]$ before use.
\subsection{Proposer Prompt}

\begin{quote}\small
\textbf{Task:} Create a challenging and self-contained reasoning task suitable for evaluating general intelligence or instruction following.

\textbf{Requirements:}
\begin{itemize}
    \item The task must require multi-step reasoning or structured planning.
    \item The task must be deterministic or tightly constrained.
    \item Avoid trivia, subjective prompts, web-dependent knowledge, or unsolvable questions.
    \item Output format:
\end{itemize}

\vspace{-0.5em}
\begin{verbatim}
<question>
[generated question]
</question>
\end{verbatim}
\end{quote}

\subsection{Validator Prompt}

\begin{quote}\small
\textbf{Task:} Evaluate the quality of a generated question using a strict rubric.

\textbf{Scoring Criteria:}
\begin{itemize}
    \item Unsolvable, contradictory, unsafe, or incoherent questions receive scores in $[1,3]$.
    \item Ambiguous or incomplete questions receive scores in $[4,7]$.
    \item Clear, feasible, self-contained, and logically sound questions receive scores in $[8,10]$.
\end{itemize}

The Validator first generates internal reasoning inside \texttt{<think>} tags, then outputs a final integer score inside \texttt{<score>} tags.
\end{quote}

\subsection{Solver Prompt}

\begin{quote}\small
\textbf{Task:} Produce a complete solution to the provided reasoning or planning problem.

\textbf{Instructions:}
\begin{itemize}
    \item Reason step-by-step before answering.
    \item Satisfy all constraints in the task.
    \item Return the final response inside:
\end{itemize}

\vspace{-0.5em}
\begin{verbatim}
<answer>
[solution]
</answer>
\end{verbatim}
\end{quote}

\subsection{Judge Prompt}

\begin{quote}\small
\textbf{Task:} Evaluate the Solver response using a strict correctness rubric.

\textbf{Scoring Criteria:}
\begin{itemize}
    \item Any factual, logical, arithmetic, or commonsense error results in a score in $[1,3]$.
    \item Incomplete but partially correct answers receive scores in $[4,7]$.
    \item Fully correct, concise, and instruction-following answers receive scores in $[8,10]$.
\end{itemize}

The Judge produces internal reasoning inside \texttt{<think>} tags followed by a final score inside \texttt{<score>} tags.
\end{quote}

\section{Evaluation Prompts}
\label{app:eval_prompts}

This appendix summarizes the LLM-as-a-judge prompts used during benchmark evaluation. All evaluation prompts are executed using \texttt{gpt-4.1-nano}, with prompt selection determined by the benchmark-specific metric type stored in \texttt{extra\_info.metric}.

\subsection{Math Accuracy}

Used for GSM8K, MATH, AMC, Minerva, and Olympiad-style benchmarks.

\begin{quote}\small
\textbf{Task:} Determine whether the model's mathematical answer matches the ground-truth solution.

The evaluator compares:
\begin{itemize}
    \item Final numerical equivalence
    \item Mathematical correctness of reasoning
    \item Whether differently formatted answers represent the same solution
\end{itemize}

The judge first produces hidden reasoning inside \texttt{<think>} tags and then outputs:
\vspace{-0.5em}
\begin{verbatim}
<answer>TRUE</answer>
\end{verbatim}
or
\begin{verbatim}
<answer>FALSE</answer>
\end{verbatim}
\end{quote}

\subsection{Multiple-Choice Accuracy}

Used for MMLU, GPQA, ARC-C, CSQA, OBQA.

\begin{quote}\small
\textbf{Task:} Determine whether the model selected the correct answer option.

The evaluator checks:
\begin{itemize}
    \item Whether the predicted choice letter exactly matches the ground truth
    \item Whether the model explicitly provided a valid choice letter
\end{itemize}

The judge first generates internal analysis inside \texttt{<think>} tags and then outputs either:
\vspace{-0.5em}
\begin{verbatim}
<answer>TRUE</answer>
\end{verbatim}
or
\begin{verbatim}
<answer>FALSE</answer>
\end{verbatim}
\end{quote}

\subsection{Truthfulness Accuracy}

\begin{quote}\small
\textbf{Task:} Evaluate whether the generated response is factually truthful and aligned with the reference answer.

The evaluator checks:
\begin{itemize}
    \item Factual accuracy
    \item Agreement with the ground-truth response
    \item Presence of misleading or false statements
\end{itemize}

The judge produces hidden reasoning inside \texttt{<think>} tags followed by:
\vspace{-0.5em}
\begin{verbatim}
<answer>TRUE</answer>
\end{verbatim}
or
\begin{verbatim}
<answer>FALSE</answer>
\end{verbatim}
\end{quote}

\section{Notation}
\label{app:notation}

\begin{table}[t]
\centering
\small
\begin{tabular}{ll}
\toprule
Symbol & Meaning \\
\midrule
$q$ & generated question \\
$a$ & Solver answer to question $q$ \\
$v(q)$ & Validator quality score for $q$ \\
$p(q,a)$ & Judge correctness score for answer $a$ \\
$p(q)$ & average solve rate for question $q$ \\
$c_V(q)$ & Validator confidence for $q$ \\
$c_J(q,a)$ & Judge confidence for answer $a$ \\
$w_P(q)$ & PPO weight for Proposer update \\
$w_S(q,a)$ & PPO weight for Solver update \\
$f_P(q), f_S(a)$ & format-validity indicators \\
\bottomrule
\end{tabular}
\caption{Main notation used in COSE.}
\label{tab:notation}
\end{table}
\section{Artifact and License Information}
\label{app:artifacts}

We use publicly available model backbones, baselines, and evaluation benchmarks for research purposes. The Qwen and Llama model families are used according to their released licenses and terms of use. Public reasoning, mathematics, and code benchmarks are used only for held-out evaluation, and are not included in COSE's self-generated training data. Baseline methods and evaluation tools, including EvalPlus for code evaluation, are used according to their public releases and cited in the main text. Our released code and generated training artifacts will include license information and documentation describing intended research use.
\section{Compute and Infrastructure}
\label{app:compute}

COSE training was implemented using the VeRL framework with vLLM for efficient rollout generation. Depending on the backbone size, each training run used either 4 or 8 NVIDIA A100 80GB GPUs and ran for approximately two days. Smaller backbones were trained with 4 A100 GPUs, while larger backbones used 8 A100 GPUs. We evaluate open-weight backbones ranging from 0.6B to 4B parameters, as described in Section~\ref{sec:exp_setup}.

In addition to GPU training cost, our evaluation uses API-based LLM grading for benchmarks that do not have deterministic exact-match or execution-based evaluation. We use a fixed grading prompt and the same API model for all compared methods to ensure consistency. The total API cost depends on the number of evaluated checkpoints, methods, and benchmark instances. In our experiments, the API cost was used only for evaluation and not for COSE training.
\end{document}